\documentclass[journal]{IEEEtran}
\usepackage{amsmath}
\usepackage{graphicx}
\usepackage{threeparttable}
\usepackage{multirow}
\usepackage{colortbl}
\ifCLASSINFOpdf
\else
\fi
\begin{document}

\title{Gland Instance Segmentation by \\Deep Multichannel Neural Networks}
%
%
%

\author{Yan Xu, Yang Li, Mingyuan Liu, Yipei Wang, Yubo Fan, Maode Lai, and Eric I-Chao Chang*
\thanks{This work is supported by Microsoft Research under the eHealth program, the Beijing National Science Foundation in China under Grant 4152033, the Technology and Innovation Commission of Shenzhen in China under Grant shenfagai2016-627, Beijing Young Talent Project in China, the Fundamental Research Funds for the Central Universities of China under Grant SKLSDE-2015ZX-27 from the State Key Laboratory of Software Development Environment in Beihang University in China. \emph{Asterisk indicates corresponding author.}}
\thanks{Yan Xu, Yang Li, Mingyuan Liu, Yipei Wang and Yubo Fan are with the State Key Laboratory of Software Development Environment and Key Laboratory of Biomechanics and Mechanobiology of Ministry of Education and Research Institute of Beihang University in Shenzhen, Beihang University, Beijing 100191, China (email: xuyan04@gmail.com; yangli951102@gmail.com; liumingyuan95@gmail.com; edithwang525@gmail.com; yubofan@buaa.edu.cn).}
\thanks{Yan Xu is with the Microsoft Research, Beijing 100080, China (email: echang@microsoft.com).}
\thanks{Maode Lai is with the Department of Pathology, School of Medicine, Zhejiang University, China (email: lmd@zju.edu.cn).}
\thanks{*Eric I-Chao Chang is with the Microsoft Research, Beijing 100080, China (email: echang@microsoft.com).}}

\maketitle

\begin{abstract}
In this paper, we propose a new image instance segmentation method that segments individual glands (instances) in colon histology images.
This is a task called instance segmentation that has recently become increasingly important.
The problem is challenging since not only do the glands need to be segmented from the complex background, they are also required to be individually identified.
Here we leverage the idea of image-to-image prediction in recent deep learning by building a framework that automatically exploits and fuses complex multichannel information, regional, location and boundary patterns in gland histology images.
Our proposed system, deep multichannel framework, alleviates heavy feature design due to the use of convolutional neural networks and is able to meet multifarious requirement by altering channels.
Compared to methods reported in the 2015 MICCAI Gland Segmentation Challenge and other currently prevalent methods of instance segmentation, we observe state-of-the-art results based on a number of evaluation metrics.
\end{abstract}


\IEEEpeerreviewmaketitle

\section{Introduction}
\IEEEPARstart
{T}{he} latest advantages in deep learning technologies has led to explosive growth in machine learning and computer vision for building systems that have shown significant improvements in a huge range of applications such as image classification \cite{krizhevsky2012imagenet}, \cite{vggnet} and object detection  \cite{girshick2015fast}. The fully convolutional neural networks (FCN) \cite{long2015fully} permit end-to-end training and testing for image labeling; holistically-nested edge detector (HED) \cite{xie15hed} learns hierarchically embedded multi-scale edge fields to account for the low-, mid-, and high- level information for contours and object boundaries; Faster R-CNN \cite{ren2015faster} is a state-of-the-art object detection method depending on region proposal algorithms to predict object locations. FCN performs image-to-image training and testing, a factor that has become crucial in attaining a powerful modeling and computational capability of complex natural images and scenes.

\begin{figure}[!t]
\centering
\includegraphics[width=3.4in]{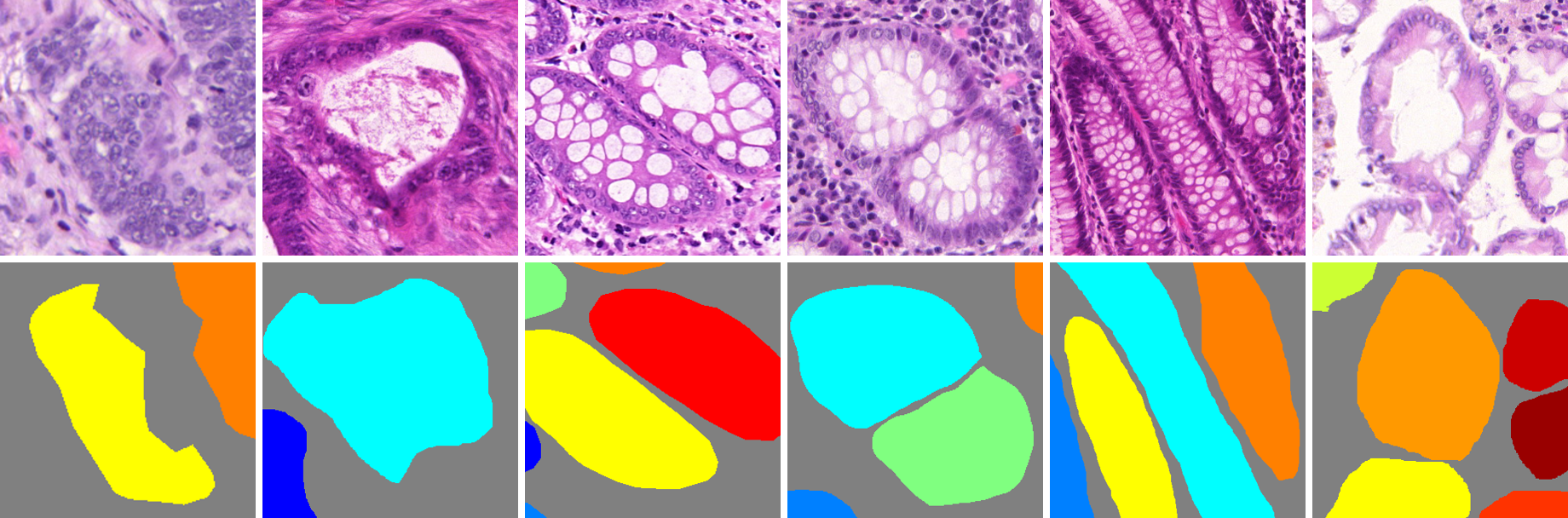}
\caption{Gland Haematoxylin and Eosin (H\&E) stained slides and ground truth labels. Images in the first row exemplify different glandular structures. Characteristics such as heterogeneousness and anisochromasia can be observed in the image. The second row shows the ground truth. To achieve better visual effects, each color represents an individual glandular structure.}
\end{figure}


The well-suited solution to image labeling/segmentation in which each pixel is assigned a label from a pre-specified set are FCN family models \cite{long2015fully,xie15hed}. However, when it concerns the problem where individual objects need to be identified, they fails to be directly applied to. This is a problem called instance segmentation. In image labeling, two different objects are assigned with the same label as long as they belong to the same class; while in instance segmentation, in addition to obtaining their class labels, it is also demanded that objects belonging to the same class are identified individually.

Exited in most organ systems as important structures, glands fulfill the responsibility of secreting proteins and carbohydrates. However, adenocarcinomas, the most prevalent type of cancer, arises form glandular epithelium. The precise instance segmentation of glands in histopathological images is essential for morphology assessment, which has been proven to be not only a valuable tool for clinical diagnosis but also the prerequisite of cancer differentiation. Nonetheless, the task of segmenting gland instances is very challenging due to the striking dissimilarity of glandular morphology in different histologic grades.


In computer vision, in spite of the promising results for instance segmentation that a recently developed progress \cite{dai2015instance} shows, it is suited for segmenting individual objects in natural scenes. With the proposal of fully convolutional network (FCN) \cite{long2015fully}, the "end-to-end" learning strategy has strongly simplified the training and testing process and achieved state-of-the-art results in solving the segmentation problem back at the time. Krahenbuh \emph{et al.} \cite{krahenbuhl2012efficient} and Zheng \emph{et al.} \cite{zheng2015conditional} integrate Conditional Random Fields (CRF) with FCN to achieve finer partitioning result of FCN. However, their inability of distinguishing different objects leads to the failure in instance segmentation problem.

The attempt to partition the image into semantically meaningful parts while classifying each part into one of pre-determined classes is called semantic segmentation and has already been well studied in computer vision. One limitation of semantic segmentation is its inability of detecting and delineating different instances of the same class while segmentation at the instance level being an important task in medical image analysis. The quantitative morphology evaluation as well as cancer grading and staging requires the instance segmentation of the gland histopathological slide images \cite{kainz2015semantic}. 
Current semantic segmentation method cannot meet the demand of medical image analysis.

The intrinsic properties of medical image pose plenty of challenges in instance segmentation \cite{dimopoulos2014accurate}. First of all, objects being in heterogeneous shapes make it difficult to use mathematical shape models to achieve the segmentation. As Fig.1 shows, the cytoplasm being filled with mucinogen granule causes the nucleus being extruded into a flat shape whereas the nucleus appears as a round or oval body after secreting. Second, variability of intra- and extra- cellular matrix is often the culprit leading to anisochromasia. Therefore, the background portion of medical images contains more noise like intensity gradients, compared to natural images.  Several problems arose in our exploration of analyzing medical image: 1) some objects lay near the others thus one can only see the tiny gaps between them when zooming in the image on a particular area; or 2) one entity borders another letting their edges adhesive with each other. We call this an issue of \emph{'coalescence'}. If these issues are omitted during the training phase, even there is only one pixel coalescing with another then segmentation would be a total disaster.

In this paper, we aim to developing a practical system for instance segmentation in gland histology images. We make use of multichannel learning, region, boundary and location cues using convolutional neural networks with deep supervision, and solve the instance segmentation issue in the gland histology image. Our algorithm is evaluated on the dataset provided by MICCAI 2015 Gland Segmentation Challenge Contest \cite{sirinukunwattana2016gland, sirinukunwattana2015stochastic} and achieves state-of-the-art performance among all participants and other popular methods of instance segmentation. We conduct a series of experiments in comparison with other algorithms and proves the superiority of the proposed framework.

This paper is arranged as follows. In section \ref{related}, a review of previous work in relative area is presented. In section \ref{method}, the complete methodology of the proposed framework of gland instance segmentation is described. In section \ref{exp}, a detailed evaluation on this method is demonstrated. In section \ref{con}, we give our conclusion.

\section{Related Work}
\label{related}
In this section, a retrospective introduction about instance segmentation will be delivered. Then, to present related work about our framework as clear as possible, information about channels will be delivered respectively, preceded by an overall review of multi-channel framework.
\subsection{Instance segmentation}
Instance segmentation, a task requires distinguishing contour, location, class and the number of objects in image, is attracting more and more attentions in image processing and computer vision. As a complex problem hardly be solved by traditional algorithms, more deep learning approaches are engaged to solve it. For example, SDS \cite{hariharan2014simultaneous} precedes with a proposal generator and then two parallel pathways for region foreground and bounding box are combined as outcome of instance segmentation. Hypercolumn \cite{hariharan2015hypercolumns}, complishes instance segmentation by utilizing hypercolumn features instead of traditional feature maps. MCNs \cite{dai2015instance} category predicted pixels via the result of object detection. In DeepMask \cite{pinheiro2015learning} and SharpMask \cite{pinheiro2016learning} two branches for segmentation and object score are engaged. Different form DeepMask, InstanceFCN \cite{dai2016instance} exploits local coherence rather than high-dimensional features to confirm instances. DCAN \cite{chen2016dcan}, the winner of 2015 MICCAI who shares the same dataset with us, combine contour and region for instance segmentation. To sum up, most of the models mentioned above make contributions to instance segmentation by integrating more than one CNN models to provide proposals and do segmentation.
\subsection{Multichannel Model}
Inspired by models mentioned above, since more than one model should engaged to solve instance segmentation problem, in another words, more than one kind of information is required, then building up a multichannel framework is also a plausible method. Multichannel model usually utilized to integrate features of various kinds to achieve more satisfying result. As far as we know, multichannel framework is rarely seen in instance segmentation of medical images. It can be seen in grouping features \cite{lee2014multi}, face recognizing \cite{chen2015learning} and image segmentation \cite{scharwachter2013efficient}, which leverage a bag-of-feature pipeline to improve the performance. In our multichannel framework, three channels aim at segmentation, object detection and edge detection are fused together. Related work about them are introduced respectively as follows.

\subsubsection{Image Segmentation}
Image segmentation aim at producing pixelwise labels to images. In neural network solution, the fully convolutional neural network \cite{long2015fully} takes the role of a watershed. Before that, patchwise training is common. Ciresan \emph{et al.} \cite{ciresan2012deep} utilize DNN to segment images of electron microscope. Farabet \emph{et al.} \cite{farabet2013learning} segment natural scene and label them. Liu \emph{et al.} \cite{liu2015crf} extract features of different patches from superpixel. Then CRF is trained to provide ultimate segmentation result. FCN \cite{long2015fully} puts forward a more efficient model to train end-to-end. After that, fully convolutional network attracts people’s attention. U-net \cite{ronneberger2015u} preserve more context information by maintaining more feature channels at up-sampling part compared to FCN. Dai \emph{et al.} \cite{dai2016instance} improve FCN model to solve instance segmentation problems.

We leverage the FCN model to produce the information of probability masks, as the region channel in our framework.
\subsubsection{Object Detection}
Object detection problem requires a system locate objects of different classes within an image. A common deep convolutional neural network solution is usually running a classier on candidate proposals and many models have been arose and improved on the basis of it. R-CNN \cite{girshick2014rich}  is a representative approach that proposals are generated by an unsupervised algorithm and classified by SVMs by features extracted by DNN. To accelerate R-CNN, fast R-CNN \cite{girshick2015fast} and faster R-CNN \cite{ren2015faster} are put forward one after another. Different from R-CNN, DeepMultiBox \cite{erhan2014scalable} generate proposals by using DNN. The network for proposal producing in OverFeat \cite{sermanet2013overfeat} share weights with network designed for classification tasks and reconcile results of classification and proposals as the ultimate result. YOLO \cite{redmon2015you} is an end-to-end model that predict proposals and class probabilities simultaneously by regard object detection problem as a single regression problem.
\subsubsection{Edge detection}
Approaches to computational edge detection play a fundamental role in the history of image processing. In recent years, solutions of neural network, at once flourishing and effective, bring about a new access towards solving complicated edge detection problems. HED \cite{xie15hed} earns hierarchically embedded multi-scale edge fields to account for the low-, mid-, and high- level information for contours and object boundaries. DeepEdge \cite{bertasius2015deepedge} also utilize multi-scale of image to solve this problem by using deep convolutional neural network. Ganin \emph{et al.} \cite{ganin2014n} propose an approach to solve edge detection problem by integrate the neural network with the nearest neighbor search. Shen \emph{et al.} \cite{shen2015deepcontour} propose a new loss function and improve the accuracy of counter detection via a neural network. 
\subsection{Previous work}
Earlier conference version of our approach were presented in Xu \emph{et al.}\cite{xu2016gland}. Here we further illustrate that: (1) we add another channel - detection channel - in this paper, due to the reason that the region channel and the detection channel complement each other; (2) this framework achieves state-of-the-art results; (3) to address the problem of images with rotation invariance, we find a new data augmentation strategy that is proven to be effective. (4) ablation experiments are carried out to corroborate the effectiveness of the proposed framework.
\section{Method}
\label{method}
\begin{figure}[!t]
\centering
\includegraphics[width=3.4in]{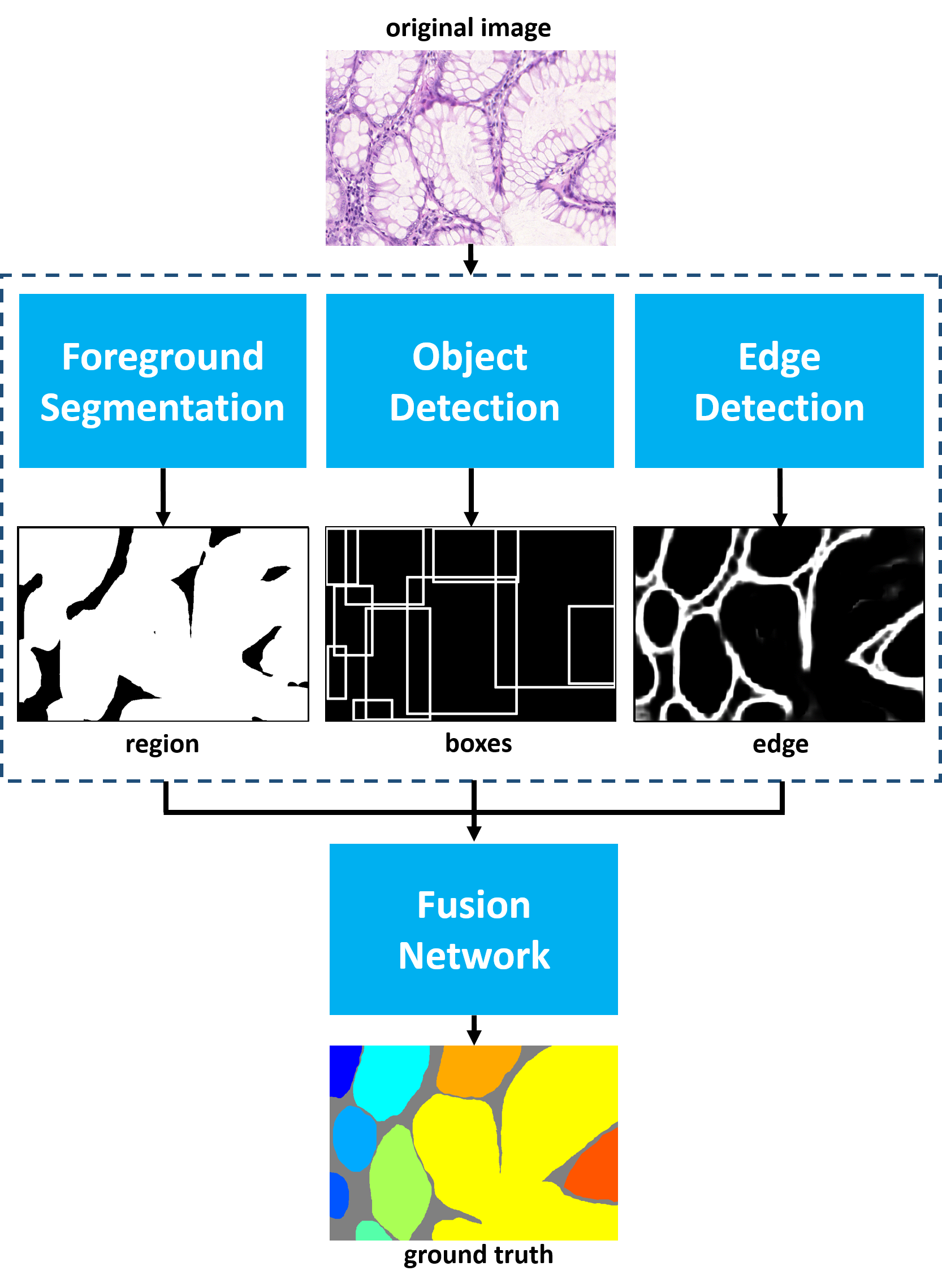}
\caption{This illustrates a brief structure of the proposed framework. The foreground segmentation channel distinguishes glands from the background. The object detection channel detects glands and their region in the image. The edge detection channel outputs the result of boundary detection. A convolution neural network concatenates features generated by different channels and produces segmented instances.}
\label{model1}
\end{figure}

In this section, we will introduce details about our framework (as shown in Fig.~\ref{model1}).

By integrating the information generated from different channels, our multichannel framework is capable of instance segmentation. Aiming at solving this problem, we select three channels, foreground segmentation channel for image segmentation, object detection channel for gland detection and edge detection channel. The reason of choosing these three channels is based on the fact that information of region, contour and location contributes receptively and complimentarily to our ultimate purpose and the joint effort of them will perform much better than each of them alone. In our framework, effects of different channels are distinct. The foreground segmentation channel in our framework distinguishes the foreground and background of images. Targeted regions in pathological contains complex morphological features. It is common that glands grow close to one another. This, however, will bring about a negative effect for algorithm that the distance between two adjacent glands are too diminutive to distinguish by computer. Machines tend to conflate two glands as a whole even there do exist a gap between them. 
Therefore, the object detection channel is in demand of designating the bounding box of each gland to which foreground pixels in that box are belonged. 
In regard to the overlapping area of bounding boxes, glands boundaries are predicted by the edge detection channel. 
As for the area that glands are close, edge detection fails to precisely predict boundaries of glands and requires the assistance of object detection channel. 
Only under the joint effort of various kinds of information, can instance segmentation problem be properly solved.
 
\begin{figure*}[!t]
\centering
\includegraphics[width=7in]{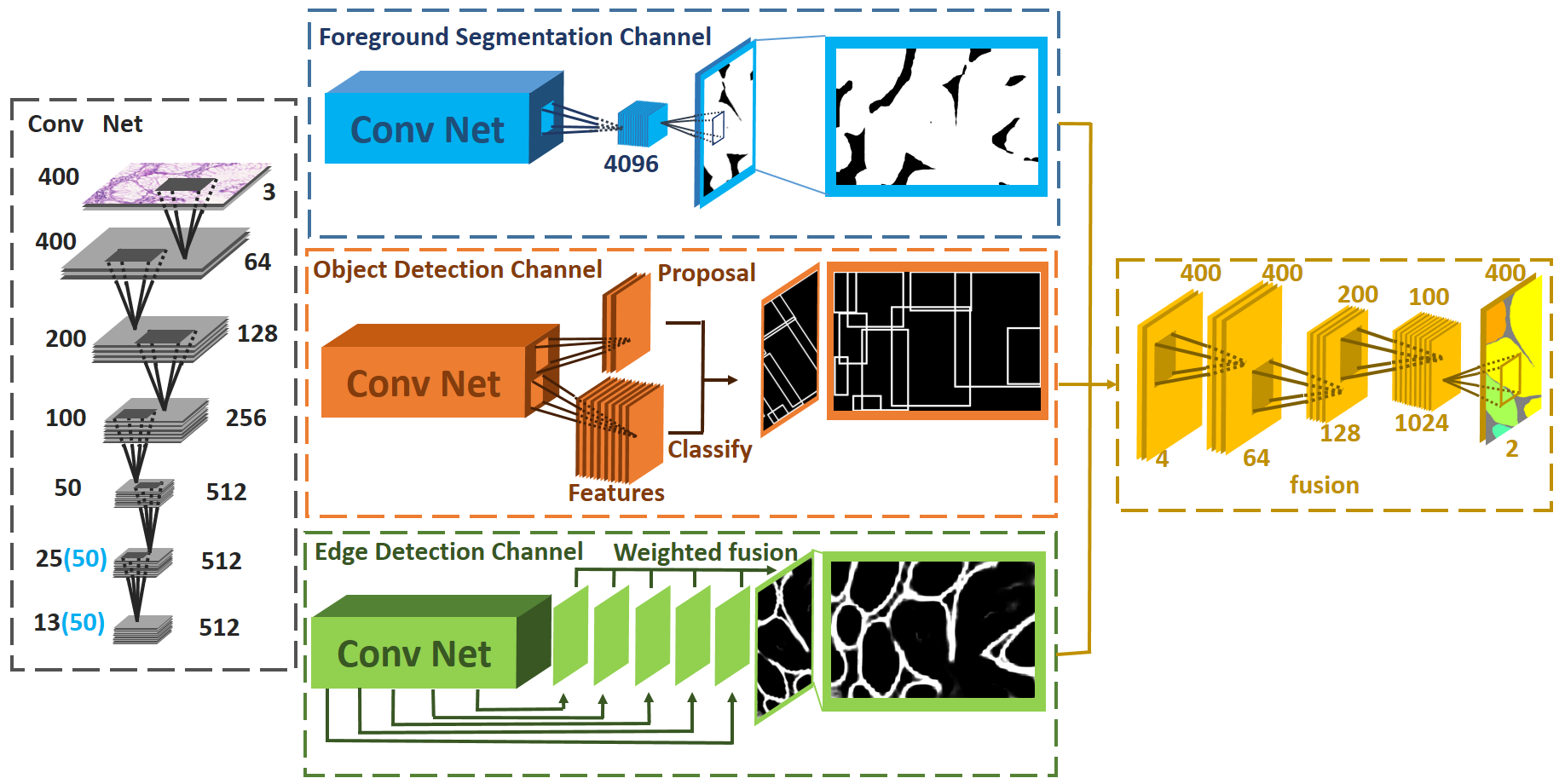}
\caption{This illustrates the structure of this framework. We fuse outputs of three channels to achieve instance segmentation. For all the channels in this framework, FCN for region channel, Faster RCNN for object channel and HED for edge channel, are all based on VGG16 model, we present this classical five pooling structure in details by "Conv Net" at the top of the figure and show it briefly by a rectangular block named "Conv Net". Especially, in region and object channels, arrows pointing from "Conv Net" denotes the output of the "Conv Net", while in edge channel they represent output of deep supervisions. In the region channel, strides of the last two pooling layers of "Conv Net" are set as 1; atrous convolution being applied to convolution layers leads to the higher resolution of feature maps (as annotated in brackets). }
\label{model2}
\end{figure*}

\subsection{Foreground Segmentation Channel}
The foreground segmentation channel distinguishes glands from the background. 

With the arising of FCN, image segmentation become more effective thanks to the end-to-end training strategy and dense prediction on attribute-sized images. FCN replace the fully-connected layer with a convolutional layer and upsample the feature map to the same size as the original image through deconvolution thus an end-to-end training and prediction is guaranteed. Compared to the previous prevalent method sliding window in image segmentation, FCN is faster and simpler. FCN family models \cite{long2015fully,xie15hed} have achieved great accomplishment in labeling images. Usually, the FCN model can be regarded as the combination of a feature extractor and a pixel-wise predictor. Pixel-wise predictor predict probability mask of segmented images. The feature extractor is able to abstract high-level features by down-sampling and convolution. Though, useful high-level features are extracted, details of images sink in the process of max-pooling and stride convolution. Consequently, when objects are adjacent to each other, FCN may consider them as one. It is natural having FCN to solve image segmentation problems. However, instance segmentation is beyond the ability of FCN. It requires a system to differentiate instances of the same class even they are extremely close to each other. Even so, the probability mask produced by FCN still performs valuable support on solving instance segmentation problems. 

To compensate the resolution reduction of the feature map due to downsampling, FCN introduce the skip architecture to combine deep, semantic information and shallow,  appearance information. Nevertheless, DeepLab \cite{chen2016deeplab} proposes the FCN with atrous convolution that empowers the network with the wider receptive field without downsampling. Less downsampling layer means less space-invariance brought by downsampling which is benefit to the enhancement of segmentation precision. 

Our foreground segmentation channel is a modified version of FCN-32s \cite{long2015fully} of which strides of pool4 and pool5 are 1 and subsequent convolution layers enlarge the receptive field by the atrous convolution. 

Given an input image $X$ and the parameter of FCN network is denoted as ${w}_{s}$, thus the output of FCN is 
\begin{equation}
{P}_{s}\left(Y^{*}_{s}=k \mid X;w_{s}\right) = {\mu}_{k}\left(h_{s}\left(X,w_{s}\right)\right),
\end{equation}
where $\mu(\cdot)$ is the softmax function. $\mu_{k}(\cdot)$ is the output of the $k$th category and $h_{s}(\cdot)$ outputs the feature map of the hidden layer.

\subsection{Object Detection Channel}
The object detection channel detects glands and their locations in the image. 

The location of object is helpful on counting number and identifying the range of objects. According to some previous works on instance segmentation, such as MNC \cite{dai2016instance}, confirmation of the bounding-box is usually the first step towards instance segmentation. After that, segmentation and other options are carried out within bounding boxes. Though this method is highly approved, the loss of context information caused by limited receptive fields and bounding-box may exacerbate the segmentation result. Consequently, we integrate the information of location to the fusion network instead of segmenting instances within bounding boxes.
To achieve the location information, Faster-RCNN, a state-of-the-art object detection model, is engaged to solve this problem. In this model, convolutional layers are proposed to extract feature maps from images. After that, Region Proposal Network (RPN) takes arbitrary-sized feature map as input and produces a set of bounding-boxes with probability of objects. Region proposals will be converted into regions of interest and classified to form the final object detection result. 

Filling is operated in consonance with other two channel and to annotate the overlapping area. Sizes of various channels should be the same before gathering into the fusion network. To guarantee the output size of object detection channel being in accordance with other channels, we reshape it and change the bounding box into another formation. The value of each pixel in region covered by the bounding box equals to the number of bounding boxes it belongs to. For example, if a pixel is in the public area of three bounding boxes, then the value of that pixel will be three. 

We denote $w_{d}$ as the parameter of Faster-RCNN and $\phi$ represents the filling operation of bounding box. The output of this channel is 
\begin{equation}
P_{d}\left(X,w_{d}\right) = \phi\left(h_{d}\left(X,w_{d}\right)\right).
\end{equation}
$h_{d}\left(\cdot\right)$ is the predicted coordinate of the bounding box.

\subsection{Edge Detection Channel}
The edge detection channel detects boundaries between glands.

The combination of merely the probability mask predicted by FCN and the location of glands tend to fuzzy boundaries of glands, especially between adjacent objects, consequently it is tough to distinguish different objects. To receive precise and clear boundaries, the information of edge is crucial which has also been proved by DCAN \cite{chen2016dcan}. The effectiveness of edge in our framework can be concluded into two aspects. Firstly, edge compensate the information loss caused by max-pooling and other operations in FCN. As a result, the contours become more precise and the morphology become more similar to the ground truth. Secondly, even if the location and the probability mask are confirmed, it is unavoidable that predicted pixel regions of adjacent objects are still connected. Edge, however, is able to differentiate them apart. 
As expected, the synergies among region, location and edge finally achieve the state-of-the art result. The edge channel in our model is based on Holistically-nested Edge Detector (HED) \cite{xie15hed}. It is a CNN-based solution towards edge detection. It learns hierarchically embedded multi-scale edge fields to account for the low-, mid-, and high- level information for contours and object boundaries. In edge detection task, pixels of labels are much less than pixels of back ground. The imbalance may decrease the convergence rate or even cause the non-convergence problem. To solve the problem, deep supervision \cite{lee2015deeply} and balancing of the loss between positive and negative classes are engaged. In total, there are five side supervisions which are established before each down-sampling layers. 

We denote $w_{e}$ as the parameter of HED, thus the $m$th prediction of deep supervision is
\begin{equation}
P^{(m)}_{e}(Y^{(m)*}_{e}=1 \mid X;w_{e})=\sigma(h^{(m)}_{e}(X,w_{e}).
\end{equation}
$\sigma(\cdot)$ denotes sigmoid function - the output layer of HED. $h^{(m)}_{e}$ represents the output of the hidden layer that relative to $m$th deep supervision. The weighted sum of M outputs of deep supervision is the final result of this channel and the weighted coefficient is $\alpha$. This process is delivered through the convolutional layer. The back propagation enables the network to learn relative importances of edge predictions under different scales.
\begin{equation}
P_{e}(Y^{*}_{e}=1 \mid X;w_{e},\alpha) = \sigma(\sum_{m=1}^{M}\alpha^{(m)}\cdot h^{(m)}_{e}(X,w_{e})).
\end{equation}

\subsection{Fusing Multichannel}
Merely receiving the information of these three channels is not the ultimate purpose of our algorithm. Instance segmentation is. As a result, a fusion system is of great importance to maximize synergies of these three kinds of information above. It is hard for a non-learning algorithm to recognize the pattern of all these information. Naturally, a CNN based solution is the best choice.

After obtaining outputs of these three channels, a shallow seven-layer convolutional neural network is used to combine the information and yield the final result. To reduce the information loss and ensure sufficiently large reception field, we once again replace downsampling with the atrous convolution.

We denote $w_{f}$ as the parameter of this network and $h_{f}$ as the hidden layer. Thus the output of the network is
\begin{equation}
P_{f}\left(Y_{f}^{*}=k\mid P_{s},P_{d},P_{e};w_{f}\right)=\mu_{k}\left(h_{f}\left(P_{s},P_{d},P_{e},w_{f}\right)\right).
\end{equation}

\section{Experiment}
\label{exp}
\begin{figure*}[!t]
\centering
\includegraphics[width=7in]{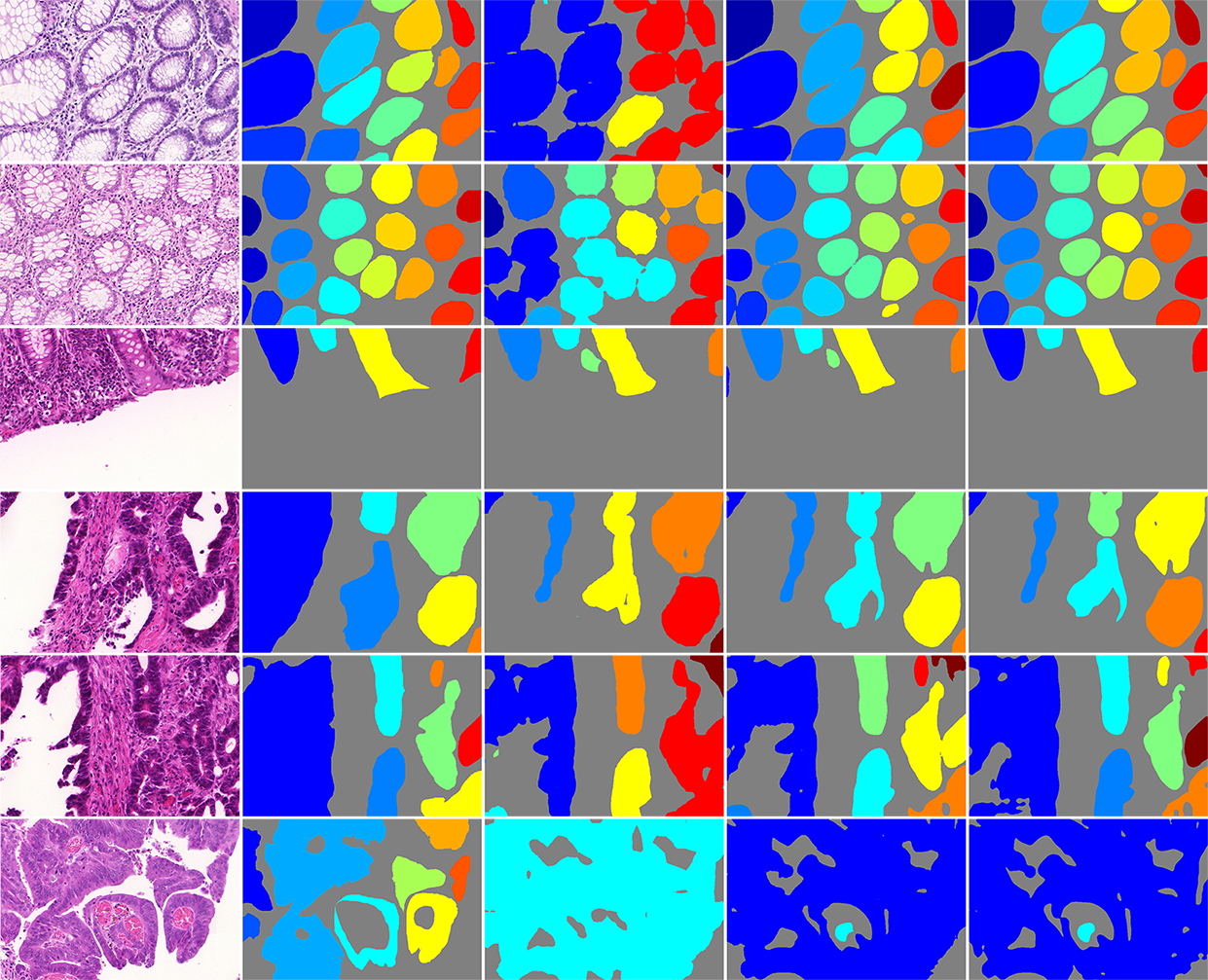}
\caption{From left to right: original image, ground truth, results of FCN, results of FCN with atrous convolution and results of the proposed framework. Compared to FCN, most of adjacent glandular structures are separated apart which indicates that our framework accomplishes the instance segmentation goal. However, few glands with small sizes or filled with red blood cells escape the detection of our model. The bad performance in the last row is due to the fact that in most samples, the white area is recognized as cytoplasm while in this sample, the white area is the background.
}
\label{result}
\end{figure*}

\begin{table*}[t]
\centering
\resizebox{\textwidth}{!}{ 
\begin{threeparttable}[b]
\caption{Performance in Comparison to Other Methods}
\begin{tabular}{c|c|c|c|c|c|c|c|c|c|c|c|c|c|c}
  \hline
   \multirow{3}{*}{Method}&
   \multicolumn{4}{c|}{F1 Score}&
   \multicolumn{4}{c|}{ObjectDice}&
   \multicolumn{4}{c|}{ObjectHausdorff}&
   \multirow{3}{*}{Rank Sum}&
   \multirow{3}{*}{Weighted Rank Sum}\\
  \cline{2-13}
   &\multicolumn{2}{c|}{Part A}&
   \multicolumn{2}{c|}{Part B}&
   \multicolumn{2}{c|}{Part A}&
   \multicolumn{2}{c|}{Part B}&
   \multicolumn{2}{c|}{Part A}&
   \multicolumn{2}{c|}{Part B}& \\
  \cline{2-13}
  & Score & Rank &  Score & Rank & Score & Rank & Score & Rank & Score & Rank & Score & Rank & \\
  \hline
  FCN & 0.788 & 11 & 0.764 & 4 & 0.813 & 11 & 0.796 & 4 & 95.054 & 11 & 146.2478 & 4 & 45 & 27.75 \\
  \hline
  FCN with atrous convolution \cite{chen2016deeplab} & 0.854 & 9 & 0.798 & 2 & 0.879 & 6 & 0.825 & 2 & 62.216 & 9 &  118.734 & 2 & 30 & 19.5 \\
  \hline\hline
  \textbf{\cellcolor[rgb]{.9,.9,.9}Ours}  & \cellcolor[rgb]{.9,.9,.9}0.893 & \cellcolor[rgb]{.9,.9,.9}3 & \textbf{\cellcolor[rgb]{.9,.9,.9}0.843} & \textbf{\cellcolor[rgb]{.9,.9,.9}1} & \textbf{\cellcolor[rgb]{.9,.9,.9}0.908} & \textbf{\cellcolor[rgb]{.9,.9,.9}1} & \textbf{\cellcolor[rgb]{.9,.9,.9}0.833} & \textbf{\cellcolor[rgb]{.9,.9,.9}1} & \textbf{\cellcolor[rgb]{.9,.9,.9}44.129} & \textbf{\cellcolor[rgb]{.9,.9,.9}1} & \textbf{\cellcolor[rgb]{.9,.9,.9}116.821} & \textbf{\cellcolor[rgb]{.9,.9,.9}1} & \cellcolor[rgb]{.9,.9,.9}8 & \cellcolor[rgb]{.9,.9,.9}4.5\\
  \hline
  CUMedVision2 \cite{chen2016dcan} & \textbf{0.912} & \textbf{1} & 0.716 & 6 & 0.897 & 2 & 0.781 & 8 & 45.418 & 2 & 160.347 & 9 & 28 & 9.5\\
  \hline
  ExB3 & 0.896 & 2 & 0.719 & 5 & 0.886 & 3 & 0.765 & 9 & 57.350 & 6 & 159.873 & 8 & 33 & 13.75\\
  \hline
  ExB2 & 0.892 & 4 & 0.686 & 9 & 0.884 & 4 & 0.754 & 10 & 54.785 & 3 & 187.442 & 11 & 41 & 15.75\\
  \hline
  ExB1 & 0.891 & 5 & 0.703 & 7 & 0.882  & 5 & 0.786 & 5 & 57.413 & 7 & 145.575 & 3 & 32 & 16.5\\
  \hline
    Frerburg2 \cite{ronneberger2015u} & 0.870 & 6 & 0.695 & 8 & 0.876 & 7 & 0.786 & 6 & 57.093 & 4 & 148.463 & 6 & 37 & 17.75\\
  \hline
  Frerburg1 \cite{ronneberger2015u} & 0.834 & 10 & 0.605 & 11 & 0.875 & 8 & 0.783 & 7 & 57.194 & 5 & 146.607 & 5 & 46 & 23\\
  \hline
  CUMedVision1 \cite{chen2016dcan} & 0.868 & 7 & 0.769 & 3 & 0.867 & 10 & 0.800 & 3 & 74.596 & 10 & 153.646 & 7 & 40 & 23.5\\
  \hline
  CVIP Dundee & 0.863 & 8 & 0.633 & 10 & 0.870 & 9 & 0.715 & 11 & 58.339 & 8 & 209.048 & 13 & 59 & 27.25\\
  \hline
  LIB & 0.777 & 12 & 0.306 & 14 & 0.781 & 12 & 0.617 & 13 & 112.706 & 13 & 190.447 & 12 & 76 & 37.5\\
  \hline
  CVML & 0.652 & 13 & 0.541 & 12 & 0.644 & 14 & 0.654 & 12 & 155.433 & 14 & 176.244 & 10 & 75 & 39.25\\
  \hline
  vision4GlaS & 0.635 & 14 & 0.527 & 13 & 0.737 & 13 & 0.610 & 14 & 107.491 & 12 & 210.105 & 14 & 80 & 39.5\\
  \hline
 \end{tabular}\end{threeparttable}}
\label{table}
\end{table*}

\subsection{Dataset}
Our method is evaluated on the dataset provided by MICCAI 2015 Gland Segmentation Challenge Contest \cite{sirinukunwattana2016gland, sirinukunwattana2015stochastic}. The dataset consists of 165 labeled colorectal cancer histological images scanned by Zeiss MIRAX MIDI. The resolution of the image is approximately 0.62μm per pixel. 85 images belong to training set and 80 affiliate to test sets (test A contains 60 images and test B contains 20 images). There are 37 benign sections and 48 malignant ones in training set, 33 benign sections and 27 malignant ones in testing set A and 4 benign sections and 16 malignant ones in testing set B.\subsection{Data augmentation and Processing}
We first preprocess data by performing per channel zero mean. The next step is to generate edge labels from region labels and perform dilation to edge labels afterward. Whether pixel is edge or not is decided by four nearest pixels (over, below, right and left) in region label. If all four pixels in region channel belongs to foreground or all of them belongs to background, then this pixel is regarded as edge. To enhance performance and combat overfitting, copious training data are needed. Given the circumstance of the absence of large dataset, data augmentation is essential before training. Two strategies of data augmentation has been carried out and the improvement of results is a strong evidence to prove the efficiency of data augmentation. In Strategy \uppercase\expandafter{\romannumeral1}, horizontal flipping and rotation operation ($0^\circ$, $90^\circ$, $180^\circ$, $270^\circ$) are used in the training images. Besides operations in Strategy \uppercase\expandafter{\romannumeral1}, Strategy \uppercase\expandafter{\romannumeral2} also includes sinusoidal transformation, pin cushion transformation and shear transformation. Deformation of original images is beneficial to the increasement of robustness and the promotion of final result. After data augmentation, a $400 \times 400$ region is cropped from the original image as input.
\subsection{Evaluation}
Evaluation method is the same as the competition goes. Three indicators are involved to evaluate performance on test A and test B. Indicators assess detection result respectively, segmentation performance and shape similarity. Final score is the summation of six rankings and the smaller the better. Since image amounts of test A and test B are of great difference, we not only calculate the rank sum as the host of MICCAI 2015 Gland Segmentation Challenge Contest demands, but we also list the weighted rank sum. The weighted rank sum is calculated as:
\begin{equation}
Weighted RS=\frac{3}{4}\sum test A Rank+\frac{1}{4}\sum test B Rank.
\end{equation}
The program for evaluation is given by MICCAI 2015 Gland Segmentation Challenge Contest \cite{sirinukunwattana2016gland, sirinukunwattana2015stochastic}.
The first criterion for evaluation reflets the accuracy of gland detection which is called F1score. The segmented glandular object of True Positive (TP) is the object that shares more than 50\% areas with the ground truth. Otherwise, the segmented area will be determined as False Positive (FP). Objects of ground truth without corresponding prediction are considered as False Negative (FN).
\begin{equation}
F1score = \frac{2\cdot Precision\cdot Recall}{Precision + Recall}
\end{equation}
\begin{equation}
Precision = \frac{TP}{TP+FP}
\end{equation}
\begin{equation}
Recall=\frac{TP}{TP+FN}
\end{equation}

Dice is the second criterion for evaluating the performance of segmentation. Dice index of the whole image is
\begin{equation}
D(G,S)=\frac{2(\mid G\cap S\mid)}{\mid G\mid +\mid S\mid},
\end{equation}
of which G represents the ground truth and S is the segmented result. However, it is not able to differentiate instances of same class. As a result, object-level dice score is employed to evaluate the segmentation result. The definition is as follows:
\begin{equation}
D_{object}(G,S)=\frac{1}{2}\left[\sum_{i=1}^{n_{S}}w_{i}D(G_{i},S_{i})+\sum_{j=1}^{n_{G}}\widetilde{w}_{j}D(\widetilde{G}_{i},\widetilde{S}_{i})\right],
\end{equation}
\begin{equation}
w_{i}=\frac{\mid S_{i}\mid}{\sum_{j=1}^{n_{S}}\mid S_{j}\mid},
\end{equation}
\begin{equation}
\widetilde{w}_{i}=\frac{\mid \widetilde{G}_{i}\mid}{\sum_{j=1}^{n_{G}}\mid \widetilde{G}_{j} \mid}.
\end{equation}
$n_{S}$ and $n_{G}$ are the number of instances in the segmented result and ground truth.

Shape similarity reflects the performance on morphology likelihood which plays a significant role in gland segmentation. Hausdorff distance is exploited to evaluate the shape similarity. To assess glands respectively, the index of Hausdorff distance deforms from the original formation:
\begin{equation}
H(G,S)=\mathrm{max}\left\{\underset{x\epsilon G}{sup}  \underset{y\epsilon S}{inf}\left\|x-y\right\|,\underset{y\epsilon S}{sup}  \underset{x\epsilon G}{inf}\left\|x-y\right\|\right\},
\end{equation}
to the object-level formation:
\begin{equation}
H_{object}(S,G)=\frac{1}{2}\left[\sum_{i=1}^{n_{s}}w_{i}H(G_{i},S_{i})+\sum_{i=1}^{n_{G}}\widetilde{w}_{i}H(\widetilde{G}_{i},\widetilde{S}_{i})\right],
\end{equation}
where
\begin{equation}
w_{i}=\frac{|S_{i}|}{\sum_{j=1}^{n_{S}}|S_{j}|},
\end{equation}
\begin{equation}
\widetilde{w}_{i}=\frac{|\widetilde{G}_{i}|}{\sum_{j=1}^{n_{G}}|\widetilde{G}_{j}|}.
\end{equation}
Similar to object-level dice index $n_{S}$ and $n_{G}$ represents instances of segmented objects and ground truth.

\subsection{Result and Discussion}Our framework performs well on datasets provided by MICCAI 2015 Gland Segmentation Challenge Contest and achieves the state-of-the-art result (as listed in Table \uppercase\expandafter{\romannumeral1}) among all participants \cite{sirinukunwattana2016gland}.  
We rearrange the scores and ranks in this table. Our method outranks FCN and other participants \cite{sirinukunwattana2016gland} based on both rank sum and weighted rank sum.

Compared to FCN and FCN with atrous convolution, our framework obtains better score which is a convincing evidence that our work is more effective in solving instance segmentation problem in histological images. Though, FCN with atrous convolution performs better than FCN, for atrous convolution process less poolings and covers larger receptive fields, our framework combines information of region, location and edge to achieve higher score in the dataset. The reason why our framework rank higher, is because most of the adjacent glandular structures have been separated apart, so that more beneficial to meet the evaluation index of instance segmentation, while in FCN and FCN with atrous convolution they are not. Results of comparision are illustrated in Fig.~\ref{result}.

\begin{table*}[t]
\centering
\caption{Comparison with instance segmentation methods}
 \begin{tabular}{c|c|c|c|c|c|c}
 \hline
 \multirow{2}{*}{Method} & 
 \multicolumn{2}{c|}{F1 Score} &
 \multicolumn{2}{c|}{ObjectDice} &
 \multicolumn{2}{c}{ObjectHausdorff} \\
 \cline{2-7}
  & Part A & Part B & Part A & Part B & Part A & Part B\\
 \hline
  HyperColumn \cite{hariharan2015hypercolumns} & 0.852 & 0.691 & 0.742 & 0.653 & 119.441 & 190.384\\
 \hline
  MNC \cite{dai2015instance} & 0.856 & 0.701 & 0.793 & 0.705 & 85.208 & 190.323\\
  \hline
  SDS \cite{hariharan2014simultaneous} & 0.545 & 0.322 & 0.647 & 0.495 & 116.833 & 229.853\\
  \hline
  BOX-$>$FCN with atrous convolution+EDGE3 & 0.807 & 0.700 & 0.790 & 0.696 & 114.230 & 197.360\\
  \hline
  \cellcolor[rgb]{.9,.9,.9}OURS & \cellcolor[rgb]{.9,.9,.9}\textbf{0.893} & \cellcolor[rgb]{.9,.9,.9}\textbf{0.843} & \cellcolor[rgb]{.9,.9,.9}\textbf{0.908} & \cellcolor[rgb]{.9,.9,.9}\textbf{0.833} & \cellcolor[rgb]{.9,.9,.9}\textbf{44.129} & \cellcolor[rgb]{.9,.9,.9}\textbf{116.821}\\
  \hline
 \end{tabular}
\end{table*}

\begin{table*}
\centering
\caption{Data Augmentation Strategy comparison}
\begin{tabular}{c|c|c|c|c|c|c|c}
 \hline
 \multirow{2}{*}{Strategy} &
 \multirow{2}{*}{Method} & 
 \multicolumn{2}{c|}{F1 Score} &
 \multicolumn{2}{c|}{ObjectDice} &
 \multicolumn{2}{c}{ObjectHausdorff} \\
 \cline{3-8}
  & & Part A & Part B & Part A & Part B & Part A & Part B\\
 \hline
 \multirow{2}{*}{Strategy \uppercase\expandafter{\romannumeral1}} &
 FCN & 0.709 & 0.708 & 0.748 & 0.779 & 129.941 & 159.639\\
 \cline{2-8}
  & FCN with atrous convolution \cite{chen2016deeplab} & 0.820 & 0.749 & 0.843 & 0.811 & 79.768 & 131.639\\
 \hline
 \multirow{2}{*}{Strategy \uppercase\expandafter{\romannumeral2}} & 
 FCN & 0.788 & 0.764 & 0.813 & 0.796 & 95.054 & 146.248\\
 \cline{2-8}
  & FCN with atrous convolution \cite{chen2016deeplab} & \textbf{0.854} & \textbf{0.798} & \textbf{0.879} & \textbf{0.825} & \textbf{62.216} & \textbf{118.734}\\
 \hline
\end{tabular}
\end{table*}

\begin{table*}[t]
\centering
\caption{Plausibility of Channels. We denote AMC as fusion network with atrous convolution and MC as fusion network without atrous convolution. EDGE1 represents that edge label are not dilated while EDGE3 signifies edge label dilated by a disk filter with radius 3. BOX detnotes the bounding box.}
 \begin{tabular}{c|c|c|c|c|c|c}
 \hline
 \multirow{2}{*}{Method} & 
 \multicolumn{2}{c|}{F1 Score} &
 \multicolumn{2}{c|}{ObjectDice} &
 \multicolumn{2}{c}{ObjectHausdorff} \\
 \cline{2-7}
  & Part A & Part B & Part A & Part B & Part A & Part B\\
 \hline
  MC: FCN + EDGE1 + BOX & 0.863 & 0.784 & 0.884 & 0.833 & 57.519 & 108.825\\
  \hline
  MC: FCN + EDGE3 + BOX & 0.886 & 0.795 & 0.901 & 0.840 & 49.578 & 100.681\\
  \hline
  MC: FCN with atrous convolution + EDGE3 + BOX & 0.890 & 0.816 & \textbf{0.905} & 0.841 & 47.081 & 107.413\\
  \hline
  \hline
  AMC: FCN + EDGE3 + BOX & \textbf{0.893} & 0.803 & 0.903 & \textbf{0.846} & 47.510 & \textbf{97.440}\\
 \hline
  AMC: FCN with atrous convolution + EDGE3 + BOX & \textbf{0.893} & \textbf{0.843} & 0.908 & 0.833 & \textbf{44.129} & 116.821\\
  \hline
  AMC: FCN with atrous convolution + EDGE1 + BOX & 0.876 & 0.824 & 0.894 & 0.826 & 50.028 & 123.881\\
   \hline
  \hline
  AMC: FCN with atrous convolution + BOX & 0.876 & 0.815 & 0.893 & 0.808 & 50.823 & 132.816\\
 \hline
  AMC: FCN with atrous convolution + EDGE3 & 0.874 & 0.816 & 0.904 & 0.832 & 46.307 & 109.174\\
  \hline
 \end{tabular}
\end{table*}

Ranks of test A are higher than test B in general due to the inconsistency of data distribution. In test A, most images are the normal ones while test B contains a majority of cancerous images which are more complicated in shape and lager in size. Hence, a larger receptive field is required in order to detect cancerous glands. However, before we exploit the atrous convolution algorithm, the downsampling layer not only gives the network larger receptive field but also make the resolution of the feature map decreases thus the segmentation result becomes worse. The  atrous convolution algorithm empower the convolutional neural network with larger receptive field with less downsampling layers. Our multichannel framework enhances the performance based on the FCN with atrous convolution by adding two channels - edge detection channel and object detection channel. 

Since the differences between background and foreground in histopathological image are small (3th row of Fig.~\ref{result}), FCN and FCN with atrous convolution sometimes predict the background pixel as gland thus raise the false positive rate. The multichannel framework abates the false positive by adding context of pixel while predicting object location. 

Compared to CUMedVision1 \cite{chen2016dcan}, CUMedVision2 \cite{chen2016dcan} add the edge information thus results of test A improve yet those of test B deteriorate. But our method improves both results of test A and test B after combine the context of edge and location. 

However, white regions in gland histopathological images are of two kinds: 1) cytoplasm; and 2) there is no cell or tissue (background). The difference between these two kinds is that cytoplasm usually appears surrounded by nuclei or other stained tissue. In the image of the last row in Fig.~\ref{result}, glands encircles white regions without cell or tissue causing that the machine mistakes them as cytoplasm. As for images of the 4th and 5th row in Fig.~\ref{result}, glands are split when cutting images, which is the reason that cytoplasm is mistaken as the background.

\textbf{Comparison with instance segmentation methods}
Currently, methods suitable for instance segmentation of images of natural scenes predict instances based on detection or proposal, such as SDS \cite{hariharan2014simultaneous}, Hypercolumn \cite{hariharan2015hypercolumns} and MNC \cite{dai2015instance}. One defect of this logic is its dependence on the precision of detection or proposal. If the object escapes the detection, it would evade the subsequent segmentation as well. Besides, the segmentation being restricted to a certain bounding box would have little access to context information hence impact the result. Under the condition of bounding boxes overlapping one another, which instance does the pixel in the overlapping region belongs to cannot be determined. The overlapping area falls into the category of the nearest gland in our experiment.

To further demonstrate the defect of the cascade architecture, we designed a baseline experiment. We first perform gland detection then segment gland instances inside bounding boxes. There is a shallow network (same as the fusion network) combines the information of foreground segmentation and edge detection to generate the final result. Configurations of all experiments are set the same as our method. Results are showed in Table \uppercase\expandafter{\romannumeral2} and prove to be less effective than the proposed framework. 

\subsection{Ablation Experiment}
\subsubsection{Data Augmentation Strategy}
To enhance performance and combat overfitting, data augmentation is essential before training. We observe through experiments that adequate transformation of gland images is beneficial to training. This is because that glands are in various shape naturally and cancerous glands are more different in morphology. Here we evaluate the effect on results of foreground segmentation channel using Strategy \uppercase\expandafter{\romannumeral1} and Strategy \uppercase\expandafter{\romannumeral2}. We present the results in Table \uppercase\expandafter{\romannumeral3}.

\subsubsection{Plausibility of Channels}
In the convolutional neural network, the main purpose of downsampling is to enlarge the receptive field yet at a cost of decreased resolution and information loss of original data. Feature maps with low resolution would increase the difficulty of upsample layer training. The representational ability of feature maps is reduced after upsampling and further lead to inferior segmentation result. Another drawback of downsampling is the space invariance it introduced while segmentation is space sensitive. The inconsistence between downsampling and image segmentation is obvious. The  atrous convolution algorithm empower the convolutional neural network with larger receptive field with less downsampling layers.

The comparison between segmentation performances of FCN with and without the atrous convolution shows the effectiveness of it in enhancing the segmentation precision. The foreground segmentation channel with the FCN with atrous convolution improves the performance of the multichannel framework. So does the fusion stage with the atrous convolution.

Pixels belonged to edge occupy a extremely small proportion of the whole image. The imbalance between edge and non-edge poses a severe threat to the network training that may lead to non convergence. Edge dilation can alleviate the imbalance in a certain way and improve the edge detection precision. 

To prove that these three channels truly improve the performance of instance segmentation, we conduct the following two baseline experiments: a) merely launch foreground segmentation channel and edge detection channel; b) merely launch foreground segmentation channel and object detection channel. The results is in favor of the three-channel framework with no surprise.

Results of experiments mentioned above are presented in Table \uppercase\expandafter{\romannumeral4}.

\section{Conclusion}
\label{con}
We propose a new framework called deep multichannel neural networks which achieves state-of-the-art results in MICCAI 2015 Gland Segmentation Challenge. The universal framework extracts features of  edge, region and location then concatenate them together to generate the result of instance segmentation.

In future work, this algorithm can be expanded in instance segmentation of medical images.


%

\section*{Acknowledgment}
This work is supported by Microsoft Research under the eHealth program, the Beijing National Science Foundation in China under Grant 4152033, the Technology and Innovation Commission of Shenzhen in China under Grant shenfagai2016-627, Beijing Young Talent Project in China, the Fundamental Research Funds for the Central Universities of China under Grant SKLSDE-2015ZX-27 from the State Key Laboratory of Software Development Environment in Beihang University in China.

\ifCLASSOPTIONcaptionsoff
  \newpage
\fi



%




\bibliographystyle{IEEEtran}
\bibliography{IEEEabrv,./Multi-cue}

%

\end{document}